\documentclass[11pt,twocolumn]{article}
\usepackage{arxiv}
\usepackage{url}
\usepackage{latexsym}
\usepackage{booktabs}
\usepackage{multirow}
\usepackage{graphicx}
\usepackage{amssymb}
\usepackage{amsmath}
\usepackage{array}
\usepackage{hyperref}
\usepackage{makecell}
\usepackage{rotating}
\usepackage{longtable}
\usepackage{authblk}
\usepackage{doi}
\usepackage{xcolor}

\newcommand{\botrule}{\bottomrule}
\renewcommand*{\arraystretch}{1.15}

\pdfoutput=1

\usepackage{times}
\usepackage{latexsym}

\usepackage[T1]{fontenc}

\usepackage[utf8]{inputenc}

\usepackage{microtype}

\usepackage{xspace}

\newcommand\distemist{\textsc{DisTEMIST}\xspace}
\newcommand{\xmen}{\textsc{xMEN}\xspace}
\newcommand{\quaero}{\textsc{Quaero}\xspace}
\newcommand{\bronco}{\textsc{Bronco}\xspace}
\newcommand{\mantragsc}{\textsc{Mantra} GSC\xspace}
\newcommand{\bigbio}{\textsc{BigBIO}\xspace}
\newcommand{\sapbert}{\textsc{SapBERT}\xspace}
\newcommand{\medmentions}{\textsc{MedMentions}\xspace}

\usepackage{multirow}
\usepackage{colortbl}
\usepackage{verbatimbox}
\usepackage{caption}
\usepackage{subcaption}

\usepackage[numbers]{natbib}

\newcommand{\biburl}[1]{\url{#1} [retrieved: Oct 17, 2023]}

\usepackage{hyperref}

\usepackage{lipsum}

\makeatletter
\newbox\sf@box
{\def\sf@one{#1}%
\def\sf@two{#2}%
\setbox\sf@box\hbox
\bgroup}%
{ \egroup
\ifx\@empty\sf@two\@empty\relax
\def\sf@two{\@empty}
\fi
\ifx\@empty\sf@one\@empty\relax
\subfloat[\sf@two]{\box\sf@box}%
\else
\subfloat[\sf@one][\sf@two]{\box\sf@box}%
\fi}
\makeatother

\usepackage{color}
\definecolor{deepblue}{rgb}{0,0,0.5}
\definecolor{deepred}{rgb}{0.6,0,0}
\definecolor{deepgreen}{rgb}{0,0.5,0}

\usepackage{pythonhighlight}

\usepackage{listings}

\newcommand{\cth}[1]{\multicolumn{1}{@{}c@{}}{#1}}

\newcommand\Autoref[1]{\@first@ref#1,@}
\def\@throw@dot#1.#2@{#1}
\def\@set@refname#1{
    \edef\@tmp{\getrefbykeydefault{#1}{anchor}{}}%
    \xdef\@tmp{\expandafter\@throw@dot\@tmp.@}%
    \ltx@IfUndefined{\@tmp autorefnameplural}%
         {\def\@refname{\@nameuse{\@tmp autorefname}s}}%
         {\def\@refname{\@nameuse{\@tmp autorefnameplural}}}%
}
\def\@first@ref#1,#2{%
  \ifx#2@\autoref{#1}\let\@nextref\@gobble
  \else%
    \@set@refname{#1}
    \@refname~\ref{#1}
    \let\@nextref\@next@ref
  \fi%
  \@nextref#2%
}
\def\@next@ref#1,#2{%
   \ifx#2@ and~\ref{#1}\let\@nextref\@gobble
   \else, \ref{#1}
   \fi%
   \@nextref#2%
}

\title{xMEN: A Modular Toolkit for Cross-Lingual Medical Entity Normalization}

\renewcommand{\shorttitle}{xMEN: A Modular Toolkit for Cross-Lingual Medical Entity Normalization}

\author[1]{Florian Borchert}
\author[1]{Ignacio Llorca}
\author[2]{Roland Roller}
\author[1]{Bert Arnrich}
\author[1]{Matthieu-P. Schapranow}

\affil[1]{Hasso Plattner Institute for Digital Engineering (HPI), University of Potsdam, Germany \authorcr E-Mail: \tt \{firstname.lastname\}@hpi.de} 
\affil[2]{German Research Center for Artificial Intelligence (DFKI), Berlin, Germany \authorcr E-Mail: \tt roland.roller@dfki.de}

\def\tablestretch{0.99}

\begin{document}

\twocolumn[\maketitle
\begin{abstract}
    \textbf{Objective:} To improve performance of medical entity normalization across many languages, especially when fewer language resources are available compared to English.\\
\textbf{Materials and Methods:} We introduce \xmen, a modular system for cross-lingual medical entity normalization, which performs well in both low- and high-resource scenarios. When synonyms in the target language are scarce for a given terminology, we leverage English aliases via cross-lingual candidate generation. For candidate ranking, we incorporate a trainable cross-encoder model if annotations for the target task are available. We also evaluate cross-encoders trained in a weakly supervised manner based on machine-translated datasets from a high resource domain. Our system is publicly available as an extensible \textsc{Python} toolkit.\\
\textbf{Results:} \xmen improves the state-of-the-art performance across a wide range of multilingual benchmark datasets. Weakly supervised cross-encoders are effective when no training data is available for the target task. Through the compatibility of \xmen with the \textsc{BigBIO} framework, it can be easily used with existing and prospective datasets.\\
\textbf{Discussion:} Our experiments show the importance of balancing the output of general-purpose candidate generators with subsequent trainable re-rankers, which we achieve through a rank regularization term in the loss function of the cross-encoder. However, error analysis reveals that multi-word expressions and other complex entities are still challenging.\\
\textbf{Conclusion:} \xmen exhibits strong performance for medical entity normalization in multiple languages, even when no labeled data and few terminology aliases for the target language are available. Its configuration system and evaluation modules enable reproducible benchmarks. Models and code are available online at the following URL: \url{https://github.com/hpi-dhc/xmen}
\end{abstract}
\medbreak
    \par\vspace{1.6em}]

\begin{figure*}[ht!]
\centering
\includegraphics[width=0.75\textwidth]{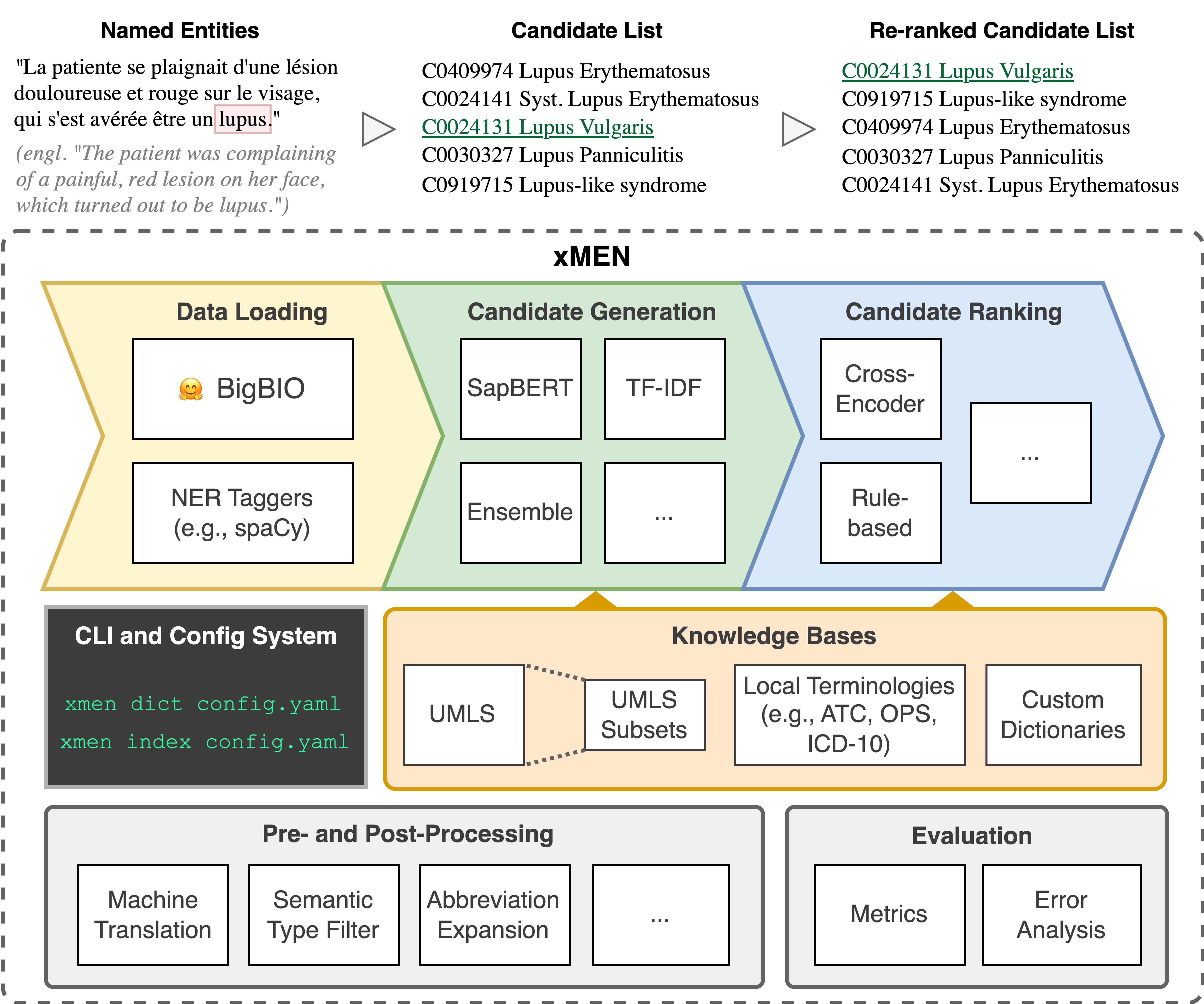}
\caption{Overview of the modular architecture of \xmen, along with a normalization example of the mention ``lupus''. \xmen can be used with any \textsc{BigBIO}-compatible dataset and implements different approaches for candidate generation and ranking, pre- and post-processing steps, evaluation metrics and utilities for error analysis. In addition, different knowledge bases can be quickly integrated and indexed via the configuration system and command line interface (CLI).}
\label{fig:overview}
\end{figure*}

\section{Introduction}

Extraction of named entities from free-text documents is a core component in most medical natural language processing (NLP) pipelines. An important sub-task is the normalization of identified entity mentions to canonical identifiers in a controlled vocabulary, ontology, or other terminology system to ensure semantic interoperability. For instance, in the Unified Medical Language System (UMLS) \cite{Bodenreider2004-pr}, the term ``lupus'' is a valid alias for ``Lupus Erythematosus'' (concept unique identifier: C0409974), ``Systemic Lupus Erythematosus (C0024141)'' and ``Lupus Vulgaris'' (C0024131). As a result, the correct disambiguation of the mention ``lupus'' will depend on the context.

At the same time, most research, software tools, and language resources for medical entity normalization (MEN) focus on the English language. However, there are particular challenges for medical texts in other languages: most concept aliases in terminology systems, like the UMLS, are only available in English, limiting the applicability of approaches that rely on such aliases to match entity mentions to concept identifiers. Moreover, most corpora annotated with grounded entity mentions necessary to train and evaluate state-of-the-art MEN approaches are also available only in English.

In this work, we present \xmen, a modular toolkit that addresses these challenges by providing modular building blocks for cross-lingual MEN. An overview of its software components is given in \autoref{fig:overview}. Following the general architecture described by \citet{Sevgili2022-gd}, we first identify potentially matching concepts in the target knowledge base (KB) for a given mention span in a candidate generation (CG) step. Second, we adapt the ranking to the specific task, also incorporating contextual information. 

\xmen is available as an extensible \textsc{Python} library. It can be used with any dataset that follows the \bigbio schema, which can be readily obtained from most span-based annotation formats and has been implemented for a wide range of MEN benchmark datasets \cite{Fries2022-my}. Our software enables users to leverage all resources available for their language, dataset, and task. When sufficient aliases in the target language are unavailable, our cross-lingual CG approach uses aliases in other languages (usually English). For subsequent task-specific re-ranking, we provide cross-encoder (CE) models in various languages. The \xmen distribution includes pre-trained models based on large-scale datasets obtained through neural machine translation (NMT). In the high-resource scenario, i.e., when sufficient annotated examples are available, we can train fully supervised CE models instead of the pre-trained ones. 

We evaluate \xmen across a diverse set of benchmark datasets in English, Spanish, French, German, and Dutch, although it is not restricted to only those languages. Leveraging the different kinds of information available for each dataset and language, \xmen matches or improves upon the state-of-the-art performance in all cases. In addition to performance improvements, we address several concerns that have been noted to impact the reproducibility of existing benchmarks \cite{Ferre2023-xx,French2023-ld}. 
These include the implementation of evaluation metrics, data pre-processing steps, and the reproducible definition of target KBs.

The contributions of our work are:
\begin{itemize}
    \item an easy-to-use and modular open-source \textsc{Python} toolkit for language-independent MEN, integrating seamlessly with existing language resources, such as the \bigbio framework and common Named Entity Recognition (NER) tools,
    \item improvement of entity normalization performance in low-resource scenarios through cross-lingual CG and weakly supervised pre-training of re-rankers, building on recent advances in NMT and label projection \cite{Chen2023-yw},
    \item a novel technique called \textit{rank regularization} to combine the ranking suggested by a general-purpose CG component with task-specific re-rankers, obtaining new state-of-the-art performance on various non-English MEN datasets, as well as
    \item a framework for reproducible benchmarks with explicit configuration of task-specific sets of target concepts and transparent evaluation criteria.
\end{itemize}

The remainder of this work is organized as follows: we set our work in the context of related work in \autoref{sec:related} followed by a description of the methods used for our \xmen toolkit in \autoref{sec:methods}. In \autoref{sec:experiments}, we share the experimental setup for our benchmarks, followed by the results in \autoref{sec:results}. We discuss the findings in \autoref{sec:discussion} and conclude our work with an outlook in \autoref{sec:conclusion}.
\section{Related Work}
\label{sec:related}

Many widely used standalone tools that implement MEN, such as  \textsc{cTAKES} \cite{Savova2010-xw}, \textsc{MetaMap} \cite{Aronson2010-ih}, \textsc{QuickUMLS} \cite{Soldaini2016-qq}, or \textsc{scispaCy} \cite{Neumann2019-lo} are based on elaborate forms of dictionary matching. Although they provide robust performance, they implicitly assume the availability of concept aliases in the target language and are usually applied to English text only. 

More recent research prototypes, such as the systems proposed by \citet{Sung2020-ip}, \citet{Bhowmik2021-tc}, \citet{Agarwal2022-fd}, or \citet{Yuan2022-sy}, achieve high-performance on benchmark datasets, such as \textsc{MedMentions} \cite{Mohan2019-xj}, \textsc{NCBI-Disease} \cite{Dogan2014-yt}, \textsc{BioCreative} V CDR \cite{Li2016-lt}, or \textsc{COMETA} \cite{Basaldella2020-rg}. For a recent survey of such systems and corpora, please refer to \citet{French2023-ld} or \citet{Garda2023-pr}. New MEN methods are frequently evaluated on datasets with sufficient annotations of grounded entity spans, which are typically only available for the English language. Moreover, these research prototypes often require complex and time-consuming setup procedures, making a comparable evaluation on other (particularly non-English) datasets challenging.

Several approaches have been proposed to address the scarcity of annotated data and concept aliases for non-English languages. Recent works leverage cross-lingual synonym relationships and other assertions from the UMLS for representation learning: \citet{Wajsburt2021-hn} present MLNorm, a system that leverages French and English synonyms for training a distantly supervised BERT (Bidirectional Encoder Representations from Transformers) classification model. The authors achieve competitive performance on the \quaero corpus \cite{Neveol2016-ww}, which can be further improved with a fully supervised model, using gold-standard annotations. To learn concept representations irrespective of the target task, \citet{Liu2021-sk} propose \textsc{SapBERT}, a BERT model that has been further fine-tuned on UMLS synonyms to better capture semantic similarity in the BERT embedding space. Using additional UMLS relations besides synonymy, \citet{Yuan2022-xa} propose CODER for learning multilingual concept representations. Their approach is evaluated across different tasks that measure semantic similarity, including MEN on the multilingual \textsc{Mantra} Gold Standard Corpus (GSC) \cite{Kors2015-hw}. In \xmen, we use such representations for unsupervised, cross-lingual CG, but combine it with a supervised re-reranking step when labelled data is available. In particular, we use the cross-lingual version of \textsc{SapBERT} \cite{liu2021learning}, which has shown competitive performance on several biomedical EL benchmarks, even without fine-tuning on task-specific data~\cite{Alekseev2022-dk}

A different approach for leveraging scarce resources is based on machine translation. \citet{Roller2018-kb} propose a multi-step system for dictionary-based candidate retrieval, combining a direct lookup using both UMLS aliases in the target language and English with a cross-lingual lookup of the machine-translated mention in the UMLS. This approach achieves competitive performance on the \quaero and \mantragsc corpora. In \xmen, we also use NMT for leveraging English-language resources. However, instead of translating mentions for the target task, we apply NMT to obtain large-scale, weakly labelled datasets in the target language and use this data to initialize a supervised, language-specific re-ranker, which can be easily shared and used across tasks. The strategy to leverage entity annotations from a high-resource language through NMT has been successfully applied for other medical information extraction tasks, such as NER \cite{Frei2022-yz,Schafer2022-od,Gaschi2023-zn}.

\section{Methods and Materials}
\label{sec:methods}

In the following, we describe the main building blocks of \textsc{xMEN} and how they can be used through the \textsc{Python} toolkit. An overview of the components of our system is depicted in \autoref{fig:overview}. 



\subsection{Data Loading}

All operations in \xmen are based on transformations of \textsc{Hugging Face} datasets built upon an extension of the \textsc{BigBIO} schema \cite{Lhoest2021-ua,Fries2022-my}. Thus, any dataset that follows this schema and provides entity annotations is compatible with \xmen. In a real-world deployment, the input for entity normalization will usually be the output of a separate NER tagger. It is straightforward to convert offset-based span annotations produced by NER taggers with \xmen. As an example, we provide an implementation for converting NER-tagged \textsc{spaCy} documents to \textsc{BigBIO}-compatible datasets as part of the package \cite{spacy}.

\subsection{Terminology Systems and Knowledge Bases}

In the general-domain entity normalization literature, the target for normalization is usually some instance of a KB, e.g., Wikipedia. In the biomedical domain, the goal is usually to link mentions to a concept identifier from a controlled vocabulary, ontology, or other medical terminology system. As MEN can be performed against any such system, we use the term \textit{knowledge base} in a task-agnostic manner to refer to the biomedical concepts of interest together with their metadata, like aliases, semantic type information, or definitions. In our experiments, target KBs are task-specific subsets of the UMLS, \textsc{Snomed} CT, as well as German versions of ICD-10 (International Statistical Classification of Diseases and Related Health Problems), OPS (Operation and Procedure Classification System), ATC (Anatomical Therapeutic Chemical Classification System), depending on the benchmark dataset. Additional aliases for KB concepts can be obtained from different sources, in particular when few aliases in the target language are available. For instance, we obtain (mostly English) aliases from \textsc{DrugBank} \cite{Wishart2017-tl} for the German version of ATC in our experiments, as described in \autoref{appendix:datasets}. 


\subsection{Pre- and Post-Processing}

For some corpora, each span is not only annotated with the gold-standard concept identifier, but also an entity type. Similarly, many NER taggers not only detect entity spans, but assign such classes. When mapping between assigned entity types and semantic type information in the target KB is possible, removing those concepts from the candidate list that are inconsistent with the given entity types is usually helpful. We have implemented components for filtering concepts based on semantic types and UMLS semantic groups in \xmen. In addition, we incorporate a simple abbreviation expansion approach, using the algorithm proposed by \citet{Schwartz2003-zt} and an implementation from \textsc{scispaCy} \cite{Neumann2019-lo}.

\subsection{Cross-lingual Candidate Generation}

Potential candidates for entity mentions in target KBs can be identified using different approaches. The encoding methods described in the following are applied to all aliases in the target KB to create an index. For inference, the same encoding is applied to mention spans, followed by an approximate nearest neighbor search to generate a ranked list of $k$ candidates. Currently, we only provide implementations of unsupervised CG approaches, which do not require training data for the target task.

\paragraph{TF-IDF with Character N-grams}

We encode all candidate aliases through character n-grams and compute TF-IDF vectors over the target KB to enable simple CG based on surface form similarity. For retrieval, we apply the same encoding to mention spans and rank candidates based on cosine similarity. Our implementation is adapted from \textsc{scispaCy}, which we extended to be compatible with non-English UMLS subsets~\cite{Neumann2019-lo}.

\paragraph{Cross-Lingual \sapbert}

We implement a CG based on the cross-lingual version of \textsc{SapBERT} to account for semantic similarity between mentions and concepts~\cite{liu2021learning}. Representations for aliases and mentions are based on the embedding of the \texttt{[CLS]} token in the last hidden layer of the BERT model. We use the cosine similarity to measure the similarity of concepts and mention embeddings. This component ensures high recall even when no aliases for a concept are available for the target language, but can be obtained from others, e.g., from the vast majority of English terms in the UMLS.

\paragraph{Ensemble}

The scored candidate lists can be combined by merging and re-sorting them according to their scores to improve overall recall. As the scores are based on cosine similarity for both CG approaches, the resulting ranking is usually informative and does not require any re-weighting \cite{Borchert2023-tu}.

\subsection{Candidate Ranking}

To adapt the task-agnostic ranking induced by unsupervised CG to specific datasets and annotation policies, we use a CE with representations similar to the approach proposed by \citet{Wu2020-zk}.
In contrast to the CG, the CE is trainable and uses the ground-truth concept labels in the training data.

\paragraph{Mention and Context Encoding}

Each mention is encoded together with its context to the left and to the right (\textit{ctx\textsubscript{l}} and {\textit{ctx\textsubscript{r}}) as follows:

\begin{center}
\texttt{[CLS]} \textit{ctx\textsubscript{l}} \texttt{[START]} \textit{mention} \texttt{[END]} \textit{ctx\textsubscript{r}}
\end{center}

with \texttt{[START]} and \texttt{[END]} denoting the beginning/end of the mention string. The context length is a configurable hyperparameter. When abbreviation expansion is performed as part of the pre-processing, we instead use the following representation:

\begin{center}
\texttt{[CLS]} \textit{ctx\textsubscript{l}} \texttt{[START]} \textit{mention} (\textit{long form}) \texttt{[END]} \textit{ctx\textsubscript{r}}
\end{center}

\paragraph{Concept Encoding}
We represent each concept by concatenating its canonical name with all its aliases (\textit{Alias\textsubscript{1..n}}), similar to the encoding proposed by \citet{Xu2020-fq}. In addition, we encode the concept's semantic type identifier to obtain the following representation:

\begin{center}
\textit{semantic type} \texttt{[TYPE]} \textit{canonical name} \texttt{[TITLE]}\\
\textit{Alias\textsubscript{1}} \texttt{[SEP]} … \texttt{[SEP]} \textit{Alias\textsubscript{n}} 
\end{center}

for example:

\begin{center}
T047 \texttt{[TYPE]} Lupus Vulgaris \texttt{[TITLE]} Lupus tuberculeux \texttt{[SEP]} Lupus exedens \texttt{[SEP]} Lupus vulgaire \texttt{[SEP]} ... \texttt{[SEP]} Tuberculosis cutis luposa 
\end{center}

The encoding can be easily adapted to include other available concept metadata, like entity descriptions \cite{Logeswaran2019-eb}. However, we did not consider these in our experiments as these are not universally available in all KBs (and only for a subset of concepts in the UMLS).

\paragraph{Training with Rank Regularization}

For training the CE, each training batch consists of $k$ candidate concepts for a single mention. As the input representation, we concatenate the mention (and context) with the concept representation. We combine a BERT-based encoder with a linear output layer to assign a score $s(m, e)$ to each such mention-concept    pair. In each batch, we include a synthetic NIL (not-in-list) concept encoded as \texttt{[UNK]} to enable the model to abstain from a prediction in cases where the correct concept is not among the candidates.

We use the CE implementation from the Sentence Transformers framework \cite{Reimers2019-ic}, and train the model equivalent to \citet{Wu2020-zk} to maximize the score of the correct candidate $e_i$ within each batch using a softmax loss:
\[ \mathcal{L}_{sm}(m, e_i) = -s(m, e_i) + log \sum_{j=1}^{k} exp(s(m_i, e_j))\]

However, the standard softmax loss neglects the potentially meaningful ranking suggested by the prior CG step. Therefore, we add a \textit{rank regularization} term to the loss function, which pushes the logits of the classification layer to preserve the original ranking and maximize top-1 accuracy. Our loss function is defined by:

\[ \mathcal{L}(m, \mathbf{e}, \mathbf{y}) = \sum_{i=1}^{k} y_i \mathcal{L}_{sm}(m, e_i) + \lambda \| s(m, \mathbf{e}) - c(m, \mathbf{e})\|_2\]

with $\mathbf{y}$ defining the vector of one-hot-encoded ground truth labels per batch, $c(m, \mathbf{e})$ is the vector of scores assigned by the CG, and $\lambda$ is a hyperparameter controlling the trade-off between top-1-accuracy and preservation of the original ranking.

\subsection{Machine Translation and Entity Alignment}

To leverage large medical corpora with grounded entity annotations in other languages as the target language, NMT methods can be used. However, to transfer the entity annotations, it is also necessary to project the original mention spans to the target language. To this end, we employ the recently proposed \textsc{EasyProject} method, an open-source tool for simultaneously performing NMT and entity alignment. It shows remarkably good performance by simply inserting special markers (like brackets \texttt{[]}) around entity mentions in the source text and translating the tagged text \cite{Chen2023-yw}. In our experiments, we use the checkpoint \texttt{ychenNLP/nllb-200-3.3b-easyproject} from the \textsc{Hugging Face} Hub, which was fine-tuned on synthetic data to preserve markers more reliably than the original NLLB (no language left behind) model \cite{NLLB_Team2022-xh}.

In our experiments, we translate \textsc{MedMentions} (ST21pv) \cite{Mohan2019-xj}, the largest English-language MEN dataset with more than 200K grounded entity mentions, to French, Spanish, German, and Dutch. The translated datasets are then used to train a supervised CE model. Since the translated and projected annotations are noisy, we refer to these models as \textit{weakly supervised} (WS). This process can be easily repeated for any language supported by \textsc{EasyProject.} 

\lstset{frame=lines,xleftmargin=2em,xrightmargin=5pt,framesep=5pt,framexleftmargin=2em,  linewidth=0.95\textwidth,numbers=left,basicstyle=\ttfamily\footnotesize,
}

\lstset{otherkeywords={%
    dict, name, umls, lang, meta_path, semantic_groups%
    },
    keywordstyle=\color{blue}
}

\hspace*{-\parindent}%
\begin{figure}[ht!]
\begin{minipage}{\hsize}%
\lstset{
language=bash,
aboveskip=0pt,
belowskip=0pt,
stringstyle=\color{stringcolour},
keywordstyle=\color{blue},
commentstyle=\color{commentcolour}\slshape,
}
\begin{lstlisting}[aboveskip=0pt,
belowskip=0pt,label=lst:yaml,caption={Example configuration for the \quaero target KB as a \textsc{Yaml} file. The configuration can be used to reproduce the subset of concepts relevant for the \quaero benchmark, where annotations are based on 10 semantic groups from the UMLS and concepts in the 2014AB release. To improve recall during candidate generation, we can obtain aliases for these concepts in different languages, here French and English.}]
name: quaero
dict:
  umls:
    lang: 
      - fr
      - en
    meta_path: ../2014AB/META
    semantic_groups:
      - ANAT
      - CHEM
      - DEVI
      - DISO
      - GEOG
      - LIVB
      - OBJC
      - PHEN
      - PHYS
      - PROC
\end{lstlisting}
\end{minipage}
\end{figure}
\lstset{frame=lines,xleftmargin=2em,xrightmargin=5pt,framesep=5pt,framexleftmargin=2em,  linewidth=0.95\textwidth,numbers=left,basicstyle=\ttfamily\footnotesize,
}
\lstset{otherkeywords={}}

\begin{figure*}[t]
\hspace*{-\parindent}%
\begin{minipage}{\hsize}%
\centering
\lstset{
language=Python,
aboveskip=0pt,
belowskip=0pt,
stringstyle=\color{stringcolour},
keywordstyle=\color{blue},
commentstyle=\color{commentcolour}\slshape,
}
\begin{lstlisting}[
label=lst:code,caption={\textsc{Python} code for candidate generation, ranking, and evaluation using the example of the \quaero dataset \cite{Neveol2016-ww}. The same pipeline can also be used for any dataset compatible with \textsc{BigBIO}. Instead of the pre-trained cross-encoder, it also possible to train a supervised model based on the training split of the dataset.}]
# Load dataset from Hugging Face Hub
import datasets
dataset = load_dataset("bigbio/quaero", "quaero_medline_bigbio_kb")

# Load knowledge base
from xmen import load_kb
kb = load_kb("path/to/quaero.jsonl")

# Candidate generation
from xmen.linkers import default_ensemble
candidate_generator = default_ensemble(index_base_path="path/to/index")
candidates = candidate_generator.predict_batch(dataset, top_k=64)

# Post-processing
from xmen.data import SemanticGroupFilter
candidates = SemanticGroupFilter(kb).transform_batch(candidates)

# Load pre-trained cross-encoder
from xmen.reranking import CrossEncoderReranker
ce_dataset = CrossEncoderReranker.prepare_data(candidates, dataset, kb)
rr = CrossEncoderReranker.load("ce_ws_medmentions_fr", device=0)

# Prediction on test set
prediction = rr.rerank_batch(candidates["test"], ce_dataset["test"])

# Evaluation  
from xmen.evaluation import evaluate
evaluate(dataset["test"], prediction)
\end{lstlisting}
\end{minipage}
\end{figure*}

\subsection{\textsc{Python} Toolkit}

All aforementioned methods are accessible through the \xmen toolkit. 
It provides a  \textsc{Python} API and a command-line interface (CLI) with \textsc{Yaml}-based  configuration  files, i.e., a human-readable markup language. \autoref{lst:yaml} shows an example of the configuration for the \quaero corpus. Given the configuration, the CLI can instantiate the UMLS subset via the \texttt{\footnotesize{xmen dict}} command. As most MEN benchmarks are based on the UMLS or its source vocabularies (like \textsc{Snomed CT}), \xmen includes an implementation for easily creating such UMLS subsets. It is also possible to pass a custom parsing script, which allows creating \xmen KBs from any source, including custom concept dictionaries or non-UMLS terminologies, such as the German OPS. For fast retrieval, indices into these KBs are pre-computed using the \texttt{\footnotesize{xmen index}}
command. 

\autoref{lst:code} depicts an example for a complete MEN pipeline for the \quaero corpus. Instead of \quaero, the same pipeline can be used for any \textsc{BigBIO}-compatible dataset with entity mention spans. More detailed usage examples can be found in the source code repository \cite{xmen-github2023}. For CG, the created indices are loaded and candidates are obtained from a \textsc{BigBIO} dataset with entity spans. The candidates can be further processed, e.g., re-ranked with any of the pre-trained models. Similarly, fully supervised re-rankers can also be trained within \xmen.

The toolkit provides implementations for the described methods for NMT, abbreviation expansion, and other pre- and post-processing techniques, which can also be used with any dataset compatible with \textsc{BigBIO}. 
It also provides fine-grained evaluation metrics by integrating the widely used \textsc{neleval} tool \cite{neleval}. Given the flexibility of \textsc{neleval}, more relaxed metrics can be easily configured for different use cases (e.g., document level scores). 

\section{Experiments}
\label{sec:experiments}

We use the gold-standard annotations in the following datasets available through the \textsc{BigBIO} framework to evaluate \xmen:
\begin{itemize}
    \item the \textsc{Mantra} gold-standard corpus (GSC) with parallel subsets in English, Spanish, French, Dutch, and German~\cite{Kors2015-hw},
    \item the French \quaero corpus~\cite{Neveol2016-ww},
    \item the German \textsc{Bronco150} corpus~\cite{Kittner2021-cm}, and
    \item the Spanish \distemist corpus~\cite{Miranda-Escalada2022-bv}.
\end{itemize}

Details regarding the text genres, annotation policy, target KBs, and the evaluation protocol for each corpus can be found in \autoref{appendix:datasets}.

For each benchmark dataset, we evaluate the performance through different CG steps and compare both the weakly and fully supervised re-reranking approaches (except for the \mantragsc as described below). All dataset-specific configurations can be found as \textsc{Yaml} files in the \xmen source code repository for reproducibility \cite{xmen-github2023}. 

We construct the target KB for each task and compute the indices for the TF-IDF- and \sapbert-based CGs. We then compute up to $k=64$ candidates for each CG and the combined candidate lists with an ensemble of both. For a fair comparison with prior work, we use entity type information in \quaero and the \mantragsc for CG \cite{Roller2018-kb,Wajsburt2021-hn}. In particular, we use a semantic type filter after the ensemble CG step to restrict the candidate set to only those UMLS concepts consistent with the gold-standard semantic groups.

For all datasets, we take the top $k$ candidates generated by the ensemble CG (plus optional semantic type filtering) and apply a weakly supervised cross-encoder \textit{CE (WS)}, pre-trained on the entire, \textsc{MedMentions} dataset translated to the respective language. This step uses no annotated training data from the target task at all. In addition, for all corpora except the \mantragsc, we train a fully supervised cross-encoder \textit{CE (FS)} on the training splits of the respective datasets. The CE models are trained for five epochs for \textsc{MedMentions} and 20 epochs for the other datasets, using a single \textsc{Nvidia} A40 GPU and keeping the checkpoint that maximizes the F$_1$ score on each dataset's validation split.

\subsection{Evaluation}

Consistent with most prior work, we evaluate systems in terms of strict (span-level) precision, recall, and $F_1$ score, as well as recall@k for different numbers of retrieved candidates $k$. Our evaluation protocol needs to account for cases, where multiple candidates for a single entity mention are provided in the gold standard. This can happen for different reasons, e.g., when the mapping from mention span to concept is ambiguous, but also when multiple concepts occur inside the same mention span. For an in-depth discussion on the issue of multi-normalization, please refer to \citet{Ferre2023-xx}. In our experiments, we treat multiple gold concepts per mention as separate linkable entities, which requires fewer assumptions about the underlying annotation policy, but is stricter than other evaluation protocols \cite{Ferre2023-xx,Garda2023-pr}.

\subsection{Model Selection and Hyperparameters}

For all adjustable parameters, we use the default settings in \xmen. For instance, we use $k=64$ candidates subject to re-ranking as suggested by \citet{Wu2020-zk}, which also coincides with the batch size we can fit into 48GB of GPU memory when training CE models. Beyond this number, recall only barely improves in our experiments. 

In addition, we have optimized the hyperparameters most relevant for training the CE through grid search using the validation sets of the \distemist and \quaero corpora. These hyperparameters are the learning rate, context length, and our newly introduced rank regularization weight $\lambda$. We do not train any models specifically for the \mantragsc, as it lacks a designated training / test split, so the corpus was not included in our grid search. 
Similarly, the evaluation for \textsc{BRONCO150} is based on 5-fold-cross-validation. Lacking a designated held-out-test set, obtaining an unbiased performance estimate after hyperparameter optimization would only be possible through nested cross-validation, which was computationally not feasible for us.

The following hyperparameters resulted in the best validation set performance:
\begin{itemize}
\setlength{\itemsep}{0pt}
\item Learning rate = $2\times10^{-5}$,
\item Context length = 128 characters, and
\item $\lambda$ = 1.0 for rank regularization.
\end{itemize}

A detailed analysis of the impact of rank regularization is presented in \autoref{sec:results}.
\section{Results}
\label{sec:results}

In the following, we share our experimental results. \autoref{tab:mantra} - \autoref{tab:distemist} summarize the benchmark results for our four benchmark datasets. We refer to individual languages by their ISO 639-2 / alpha-3 code, e.g., \textit{CE (WS\textsubscript{Fre})} refers to the weakly supervised cross-encoder trained on the French translation of \medmentions. Details on the quality of the NMT and label projection procedure can be found in \autoref{appendix:translate}.

\begin{table*}[ht!]
\renewcommand{\arraystretch}{\tablestretch}
\centering
\caption{F$_1$@1 scores for the \mantragsc through different steps of CG and after re-ranking with the weakly supervised CE for the respective language (CE (WS\textsubscript{\{Language\}}). Precision and recall have been omitted due to space constraints, as they are almost identical to the F$_1$ score for \xmen (we do not allow the weakly supervised CE to abstain from making predictions). Baselines are the NMT-based systems proposed by \citet{Roller2018-kb} as well as CODER \cite{Yuan2022-xa}. Best (highest) scores per column are highlighted \textbf{bold}, second-best \underline{underlined}.}
\label{tab:mantra}
\tabcolsep=0pt
\begin{tabular*}{0.95\textwidth}{@{\extracolsep{\fill}}lrrrrrrrrrrrrr@{\extracolsep{\fill}}}
\toprule%
& \multicolumn{5}{@{}c@{}}{MEDLINE} & \multicolumn{5}{@{}c@{}}{EMEA} & \multicolumn{3}{@{}c@{}}{Patents}
\\
\cline{2-6}\cline{7-11}\cline{12-14}%
\noalign{\vspace{0.2ex}} Language & \cth{Eng} & \cth{Spa} & \cth{Fre} & \cth{Dut} & \cth{Ger} & \cth{Eng} & \cth{Spa} & \cth{Fre} & \cth{Dut} & \cth{Ger} & \cth{Eng} & \cth{Fre} & \cth{Ger} \\
\midrule
\textbf{\xmen Candidates}  \\

TF-IDF  & .729 & .652 & .589 & .497 & .612 & .714 & .684 & .619 & .502 & .570 & .698 & .614 & .570 \\
SapBERT  & .696 & .696 & .578 & .602 & .718 & .704 & .672 & .671 & .650 & .687 & .704 & .702 & .708 \\
Ensemble  & .718 & .699 & .605 & .624 & .731 & .725 & .659 & .705 & .655 & .669 & .767 & .739 & .699 \\
Ensemble + Type Filter  & \underline{.833} & \underline{.769} & \underline{.705} & \underline{.683} & \textbf{.792} & \underline{.806} & \underline{.754} & \underline{.759} & \underline{.728} & \underline{.741} & \underline{.816} & \underline{.771} & \underline{.760}  \\
\midrule
\textbf{\xmen w/ Re-ranking} \\
CE (WS\textsubscript{\{Language\}})  & \textbf{.869} & \textbf{.838} & \textbf{.756} & \textbf{.713} & \underline{.789} & \textbf{.827} & \textbf{.789} & \textbf{.766} & \textbf{.730} & \textbf{.753} & \textbf{.857} & \textbf{.834} & \textbf{.799} \\
\midrule
\midrule
\textbf{Baseline} \\
BTM \cite{Roller2018-kb}  & \cth{--} & .691 & .674 & .614 & .663 & \cth{--} & \cth{--} & \cth{--} & \cth{--} & \cth{--} & \cth{--} & \cth{--} & \cth{--} \\
GB \cite{Roller2018-kb}  & \cth{--} & .687 & .686 & .648 & .679 & \cth{--} & \cth{--} & \cth{--} & \cth{--} & \cth{--} & \cth{--} & \cth{--} & \cth{--} \\
CODER \cite{Yuan2022-xa}  & \cth{--} & .701 & .586 & .586 & .690 & \cth{--} & .681 & .629 & .617 & .653 & \cth{--} & .708 & .690 \\
\botrule
\end{tabular*}
\end{table*}
\begin{table*}[ht!]
\renewcommand{\arraystretch}{\tablestretch}
\centering
\caption{Benchmark results with fully supervised (FS) and weakly supervised (WS) CE re-ranking for the MEDLINE and EMEA subsets of the \quaero test set (CLEF eHealth 2016). Baselines are the NMT-based system proposed by \citet{Roller2018-kb} as well as the distantly supervised (DS) and fully supervised (FS) versions of MLNorm \cite{Wajsburt2021-hn}.}
\label{tab:quaero}
\tabcolsep=0pt
\begin{tabular*}{0.75\linewidth}{
@{\extracolsep{\fill}}lrrrrrrrrr@{\extracolsep{\fill}}}
\toprule%
& \multicolumn{4}{@{}c@{}}{MEDLINE} & \multicolumn{4}{@{}c@{}}{EMEA}
\\
\cline{2-5}\cline{6-9}%
\noalign{\vspace{0.2ex}} & \cth{R@64} & \cth{P@1} & \cth{R@1}  & \cth{F$_1$@1}  & \cth{R@64}  & \cth{P@1} & \cth{R@1}  &  \cth{F$_1$@1} \\
\midrule
\textbf{\xmen Candidates}  \\
TF-IDF & .787 & .563 & .562 & .563  & .720 & .554 & .552& .553 \\
SapBERT  & \underline{.920} & .621 & .620  & .621 & .876 & .571 & .570  & .571\\
Ensemble & \textbf{.926} & .583 & .582 & .583  & \underline{.879} & .573 & .572 & .573 \\
Ensemble + Type Filter  & .918 & .663 & .661 & .662 & \textbf{.892} & .643 & .641  & .642\\
\midrule
\textbf{\xmen w/ Re-ranking} \\
CE (WS\textsubscript{Fre})  & \cth{--} & .746 & \underline{.746} & .746 & \cth{--} & .711 & \underline{.710} &  .711 \\
CE (FS)  & \cth{--} &  \underline{.795} & \textbf{.780} & \underline{.788} & \cth{--} & \underline{.801} & \textbf{.781} & \textbf{.791} \\
\midrule\midrule
\textbf{Baseline} \\
BTM \cite{Roller2018-kb} & \cth{--} & .771 & .663 & .713 & \cth{--} & .781 & .692 & .734\\
MLNorm (DS) \cite{Wajsburt2021-hn} & \cth{--}  &.775 & .734 &  .754 & \cth{--} & .746 & .709 &  .727\\
MLNorm (FS) \cite{Wajsburt2021-hn}& \cth{--}  & \textbf{.860} & .740 &  \textbf{.795}   & \cth{--}  & \textbf{.832}   & .670 & \underline{.743}\\
\botrule
\end{tabular*}
\end{table*}

\begin{table*}[ht!]
\renewcommand{\arraystretch}{\tablestretch}
\centering
\caption{Benchmark results (5-fold cross-validation) for the different entity types in BRONCO150. Our baseline is the method proposed by \citet{Kittner2021-cm}, consisting of a dictionary lookup and rule-based re-ranking.}
\label{tab:bronco}
\tabcolsep=0pt
\begin{tabular*}{0.95\textwidth}{
@{\extracolsep{\fill}}lrrrrrrrrrrrr@{\extracolsep{\fill}}}
\toprule%
& \multicolumn{4}{@{}c@{}}{Diagnosis} & \multicolumn{4}{@{}c@{}}{Treatment} & \multicolumn{4}{@{}c@{}}{Medication}
\\
\cline{2-5}\cline{6-9}\cline{10-13}%
\noalign{\vspace{0.2ex}} & \cth{R@64} & \cth{P@1} & \cth{R@1}  & \cth{F$_1$@1}  & \cth{R@64}  & \cth{P@1} & \cth{R@1}  & \cth{F$_1$@1}  & \cth{R@64}  & \cth{P@1} & \cth{R@1}  & \cth{F$_1$@1}  \\
\midrule
\textbf{\xmen Candidates}  \\

TF-IDF  & .790 & .548 & .520 & .533 & .379 & .080 & .072 & .076 & \textbf{.819} & .526 & .494  & .510 \\
SapBERT  & .890 & .639 & .638 & .639 & .544 & \underline{.210} & \underline{.192} & \underline{.201} & .784 & .481 & .452  & .466\\
Ensemble  & \textbf{.891} & \underline{.648} & \underline{.647}& \underline{.648} &  \textbf{.547} & .209 & .191 & .199 & \underline{.805} & .437 & .410 &  .423\\
\midrule
\textbf{\xmen w/ Re-ranking} \\
CE (WS\textsubscript{Ger})  & \cth{--}  & .628 & .628 & .628 & \cth{--}  & .185	& .180	& .183 & \cth{--}  & .580 & .569	& .574\\
CE (FS)  & \cth{--}  & \textbf{.807} & \textbf{.746} & \textbf{.775} & \cth{--} & \textbf{.748} & \textbf{.411} & \textbf{.530} & \cth{--} & \textbf{.807} & \textbf{.684} & \textbf{.740} \\
\midrule
\midrule
\textbf{Baseline} \\
\citet{Kittner2021-cm} & \cth{--} & .58\enspace & .54\enspace & .56\enspace & \cth{--} & .18\enspace &	.13\enspace &	 .15\enspace &  \cth{--} & \underline{.66}\enspace & \textbf{.68\enspace} & \underline{.67}\enspace \\
\botrule
\end{tabular*}
\end{table*}
\begin{table}[ht]
\renewcommand{\arraystretch}{\tablestretch}
\centering
\caption{Benchmark results on the \distemist~shared task test set of the 10th \textsc{BioASQ} workshop.}
\label{tab:distemist}
\tabcolsep=0pt
\begin{tabular*}{0.9\linewidth}{
@{\extracolsep{\fill}}lrrrr@{\extracolsep{\fill}}}
\toprule%
& \cth{R@64}  & \cth{P@1} & \cth{R@1}  & \cth{F$_1$@1} \\
\midrule
\textbf{\xmen Candidates}  \\
TF-IDF &  .767 & .332 & .319 & .325 \\
SapBERT  & \underline{.823} &.409 & .393 &  .401 \\
Ensemble  & \textbf{.830} & .435 & .418 & .426 \\
\midrule
\textbf{\xmen w/ Re-ranking} \\
CE (WS\textsubscript{Spa})  & \cth{--}  & .444 & .444  & .444 \\
CE (FS)  & \cth{--} & \textbf{.694} & \textbf{.618} & \textbf{.654} \\
\midrule
\midrule
\textbf{Baseline} \\
\citet{Borchert2023-tu} & .798 & \underline{.592} & \underline{.592} & \underline{.592} \\
\botrule
\end{tabular*}
\end{table}

\subsection{Candidate Generation Performance}


The ensemble of the TF-IDF-based and \sapbert achieves the best performance in terms of recall@64 in the majority of the cases, i.e., in the number of instances that can possibly be correctly ranked (up to 92.6\% for \quaero). The only exception is the \textit{Medications} subset of BRONCO, where the simple TF-IDF-based approach performs better by 1.4pp~(\autoref{tab:bronco}). We attribute this to the fact that medication names are usually proper nouns with little morphological variability, while the list of aliases available for CG are very comprehensive through the inclusion of \textsc{DrugBank}. Moreover, the ensemble is also competitive in terms of F$_1@1$, making it a reasonable default choice if a re-ranker is not available. For \mantragsc (\autoref{tab:mantra}) and \quaero (\autoref{tab:quaero}), where semantic type information is available, F$_1@1$ is dramatically improved by up to 11.5pp~by restricting the candidate lists to concepts with compatible semantic types. In particular, we note that the type-filtered ensemble already improves upon all baselines for \mantragsc, which can even be further improved through re-ranking.

\subsection{Weakly Supervised Re-ranking}

The pre-trained re-ranking models improve performance over the ranking suggested by the CG ensemble in most cases. For the UMLS-based annotations in the \mantragsc (\autoref{tab:mantra}) and \quaero (\autoref{tab:quaero}), the improvement is most pronounced, even on the \textit{EMEA} and \textit{Patents} subsets of both corpora, although these are different text genres than the MEDLINE abstracts comprising \medmentions. Weakly supervised re-ranking also improves performance over raw CG output for \distemist and the \textit{Medication} subset of BRONCO. This is surprising, as the target terminologies for these sub-tasks are different and do not even contain UMLS aliases in the case of BRONCO. However, for the \textit{Diagnosis} and \textit{Treatment} entities in BRONCO, the re-ranker trained on \medmentions slightly decreases performance, although only by 1--2 pp~(usually, the initial CG ranking is not altered at all in these cases). For these two scenarios, the target terminologies (ICD-10 and OPS) are comparatively small. Moreover, we use German aliases only, which is quite different from the training regime of the re-ranker. 
For both BRONCO subsets, fully supervised training on task-specific data is required, as described in the next section. 

\subsection{Fully Supervised Re-ranking}

The fully supervised re-ranker models perform best out of all \xmen configurations. Moreover, it outperforms all baselines, except for the fully supervised version of MLNorm \cite{Wajsburt2021-hn}, which obtains higher precision on both subsets of the \quaero corpus and a slightly higher $F_1$ score on the MEDLINE subset (+0.7pp). The improvements in terms of F1@1 over the strongest CG is substantial for all datasets, improving between +12.6pp on \quaero (MEDLINE) and +33.1pp on BRONCO (Treatments), achieving new state-of-the-art performance on all benchmarks except for the MEDLINE subset of \quaero.

\subsection{Impact of Rank Regularization}

During hyperparameter optimization, a value of $\lambda = 1$ performed best on the validation sets. \autoref{fig:rank_reg} shows an in-depth analysis on the impact on test set performance. It is evident that recall@1 peaks for $\lambda = 1$ also on the test sets for both \quaero and \distemist, making it a sensible choice as a default value in \xmen. However, nearby values in the range $[0.4,1.2]$ also seem to work well. For \distemist, the differences between no ($\lambda = 0$) and too much ($\lambda = 2.0$) regularization are more pronounced than for \quaero. Interestingly, the recall for larger values of $k$ slightly improves for \quaero as regularization is increased. We assume that this is due to the suppression of NIL predictions when the initial CG ranking is given higher priority over the predicted ranking.

\begin{figure}[t!]
         \centering
         \includegraphics[width=0.89\hsize]{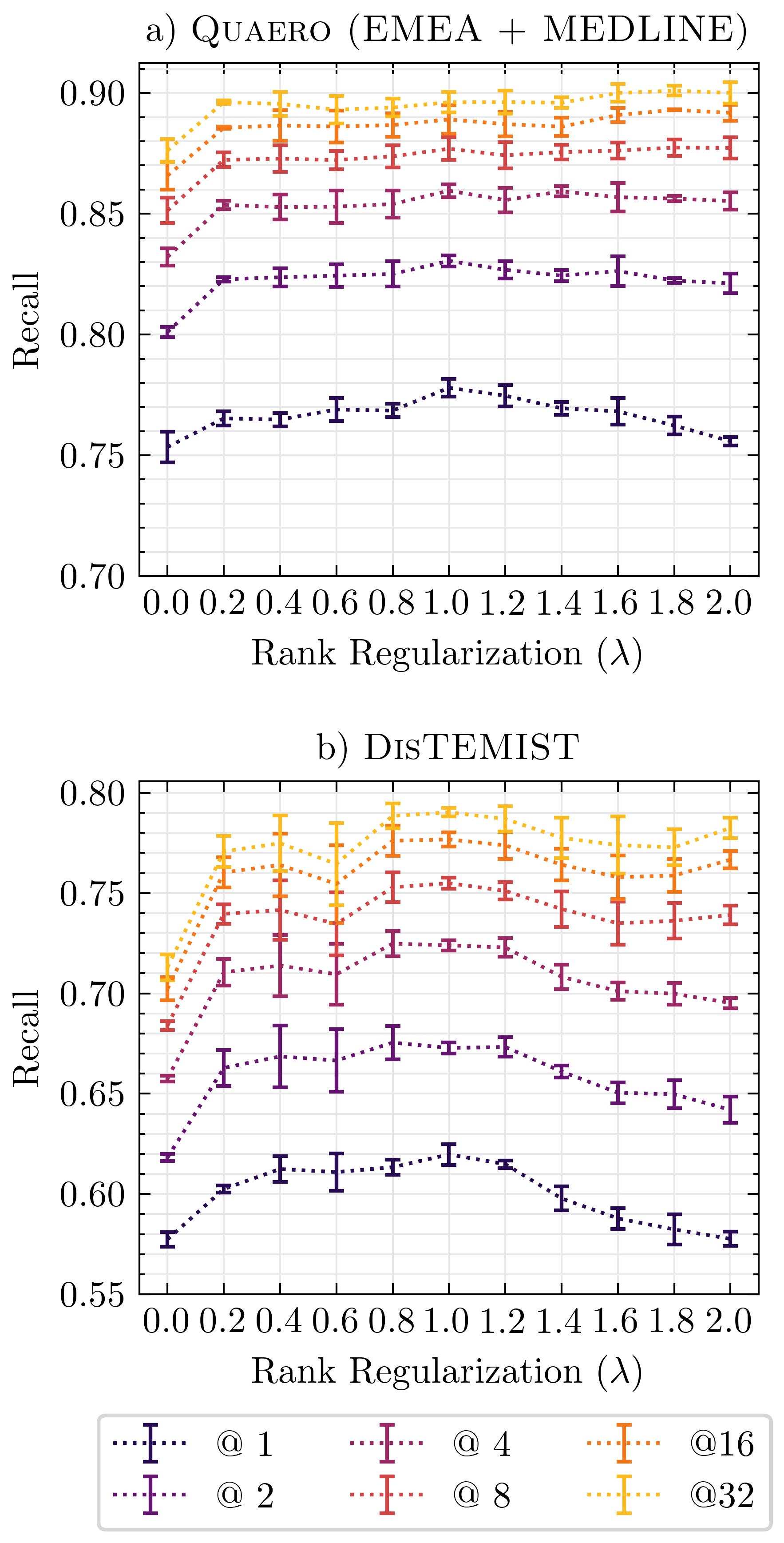}
         \caption{Impact of the relative weight $\lambda$ of the rank regularization term in the CE loss function. We report the test set recall@$k$ for different values of $k$ in a) for \quaero and b) for \distemist. For each value of $\lambda$, we report the mean and standard deviation across three runs with different random seeds. Note that the y-axes have different intervals, as the baseline performance is higher for \quaero.}
         \label{fig:rank_reg}
\end{figure}

\begin{figure*}[ht!]
         \centering
         \includegraphics[width=0.87\textwidth]{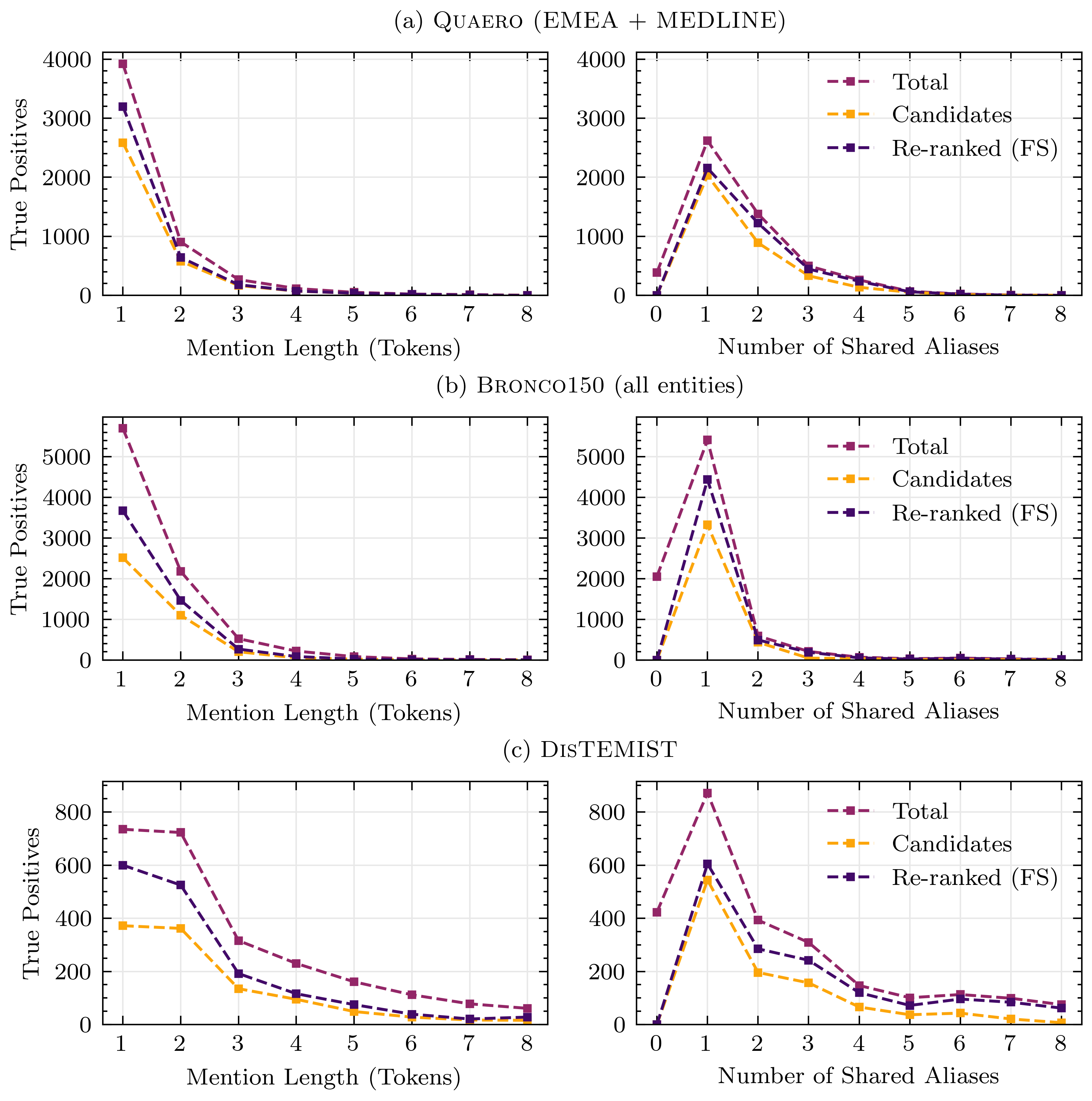}
         \caption{Impact of mention length and lexical ambiguity on the absolute number of true positives (for $k=1$) before and after re-ranking with the fully supervised CE. The ``Total'' line refers to the total number of concepts in the gold-standard. The number of shared aliases in the right column is the maximum number of aliases that any concept in the candidate lists shares with the ground truth concept. When zero aliases are shared, this means that the correct concept was not among the retrieved candidates, therefore the number of true positives is also zero, before and after re-ranking. }
         \label{fig:error_analysis}
\end{figure*}

\section{Discussion}
\label{sec:discussion}

In this section, we discuss errors of our system as well as the overall limitations of this work. We note that two conditions account for the majority of false negatives, i.e., reduced recall during CG and subsequent ranking errors: complex entity mentions (consisting of multiple tokens) and lexical ambiguity of KB aliases. In \autoref{fig:error_analysis}, we compare the (absolute) number of true positives for $k=1$, i.e., the number of correctly predicted and ranked concepts for \quaero, \bronco, and \distemist before and after fully supervised re-ranking. 

\subsection{Complex Entity Mentions}

For all analyzed corpora, recall@1 decreases for longer mention spans, both before and after re-ranking. Moreover, re-ranking can only recover from these ranking errors for shorter entity spans. For instance, the fully supervised CE trained on the \quaero training set improves recall from .659 to .816 (+15.7pp) for mentions of length one, from .639 to .710 (+7.1pp) for length two, and from .625 to .674 (+4.9pp) for length three. The results are similar for \bronco, where recall values are lower overall, but re-ranking is mostly effective for mentions of length one or two. However, the frequency of longer mention spans quickly decreases for both corpora, and most mentions are short. In comparison, \distemist contains a larger fraction of long mention spans, with low recall values for spans of length greater than two, which also hardly improve through re-ranking and hence have a large impact on the overall recall.

\subsection{Lexical Ambiguity}

In our employed CG approach, individual aliases are treated as proxies for target concepts. Consequently, concepts sharing the same alias can be identified as the nearest neighbor match and get assigned the same candidate score, such as the mention ``lupus'' in \autoref{fig:overview}, which is a valid alias for multiple UMLS concepts. 

As the target KBs for \bronco are very specific and mostly German, few aliases are usually shared among concepts, as depicted in \autoref{fig:error_analysis} (b). In most cases, the generated candidates share either zero aliases with the ground truth concept (i.e., when the concept was not retrieved as a candidate) or exactly one alias. For \quaero and \distemist, where large, multilingual KBs are employed, lexical ambiguity is a major source of ranking errors. These errors can be effectively resolved through re-ranking. In contrast, re-ranking barely impacts performance@1 when the ground truth entity shares only a single alias with the generated candidates, i.e., when the correct concept is retrieved as a candidate, but misranked for reasons other than lexical ambiguity.

\subsection{Limitations}

While we evaluated \xmen on a wide range of datasets, our work does not present an effort to provide a comprehensive, multilingual MEN benchmark. Such a benchmark should include tasks from different domains, e.g., more biologically oriented tasks, such as gene name normalization. It should also cover more (in particular non-European) languages. The employed \textsc{EasyProject} method has been evaluated on 57 languages and can be easily applied to obtain more weakly supervised re-ranking models. Moreover, given the task-agnostic implementation of data ingestion, KB configuration, and evaluation protocols in \xmen, it should be possible to integrate existing benchmarks covering more diverse datasets \cite{Garda2023-pr}. Future work should also consider more realistic training/test splits, which test the true zero-shot MEN abilities of systems \cite{Alekseev2022-dk}.

In addition, there are dimensions that could be explored to further optimize the performance of \xmen. For instance, we choose a single BERT checkpoint per language for initializing the CE models, based on reported performance on other information extraction tasks \cite{Labrak2023-qd,Lentzen2022-ia,Carrino2022-wk}. Although the chosen models work well, it is possible that differently pre-trained (e.g., multilingual or more domains-specific) models might perform better for the re-ranking problem \cite{Bressem2023-gh}.

Our error analysis identified the primary sources of CG and ranking errors. While supervised re-ranking effectively resolves lexical ambiguity, we have yet to devise a strategy for enhancing overall recall and ranking for long entity mentions. This issue has predominantly affected the \distemist corpus in our experiments, presumably due to varying annotation policies, and has been less problematic for other corpora. Moreover, a certain number of less common ranking errors cannot be attributed to long mention spans or lexical ambiguity. A more detailed investigation of these errors might allow us to implement components for \xmen to resolve them effectively.
\section{Conclusion}
\label{sec:conclusion}

In the given work, we have presented \xmen, a novel \textsc{Python} toolkit for normalizing medical entities in many languages. In particular, we combined strong unsupervised candidate generation approaches with trainable cross-encoders and a novel loss function for regularizing their training. This pipeline improves upon previous state-of-the-art performance on various benchmark datasets. We also introduce pre-trained weakly supervised re-ranking models, which can be used when little or no training data is available for the target task.

The modularity of \xmen supports a broad practical applicability. In cases where no sufficient training data is available for the target task and language is available, an ensemble of candidate generators with multilingual aliases already improves upon the state-of-the-art in numerous instances, e.g., for the \mantragsc or diagnoses and treatments in the BRONCO corpus. For UMLS-based or semantically related annotation schemes, a cross-encoder model pre-trained on the \medmentions 
dataset usually improves performance further. State-of-the-art results are achieved for all benchmarks when task-specific training data can be used to train fully supervised cross-encoder models.

There is ample opportunity to extend \xmen with new modules. For instance, it will be straightforward to implement trainable candidate generators, such as bi-encoders \cite{Wu2020-zk}, clustering-based approaches \cite{Agarwal2022-fd}, or generative models \cite{Yuan2022-sy}. Moreover, the \sapbert-based candidate generator can be easily swapped with other models, which consider other KB-information in their training procedures \cite{Yuan2022-xa,Sakhovskiy2023-nd}. Our error analysis has shown that long mention spans are particularly challenging to normalize, which might be alleviated by further pre-processing components that can be implemented within the \xmen framework.

In the future, we plan to evaluate \xmen on further English and non-English datasets once they become available to support MEN research. We believe that our implementation of a configuration system for medical terminology subsets and evaluation metrics can form the basis for a comprehensive MEN benchmark, covering a diverse set of languages, text genres and entity classes.

\section*{Acknowledgements} Parts of this work were generously supported by a grant of the German Federal Ministry of Research and Education (01ZZ2314N).
\vfill
\bibliographystyle{unsrtnat}
\bibliography{paperpile,refs}

\appendix
\newpage
\section{Benchmark Datasets}
\label{appendix:datasets}
In the following, we describe the benchmark datasets and derived target knowledge bases in  detail.

\subsection*{\mantragsc}

The multilingual \textsc{Mantra} Gold Standard Corpus comprises parallel text segments in five languages: English, Spanish, French, Dutch, and German \cite{Kors2015-hw}. 100 bilingual MEDLINE titles are available for English and each of the other languages, 100 sentences from drug labels in all five languages, as well as 50 parallel sentences from patents in English, French and German. Entities have been annotated according to the \textsc{Mantra} terminology, a UMLS subset restricted to ten semantic groups and the source vocabularies MeSH, \textsc{Snomed CT}, and MedDRA. The \mantragsc contains 5,530 annotations, but much less in individual languages (e.g., only 1,052 for the French subset). As \mantragsc also does not provide pre-defined train/validation/test splits, we do not train any fully supervised models for this dataset, but only evaluate the unsupervised CG and pre-trained CE models for each language.

\subsection*{\quaero}

The corpus was used in the CLEF eHealth 2016 lab (task 2) and consists of two subsets: 2,498 scientific articles from \textsc{Medline} and 10 drug monographs from the European Medicines Agency (EMEA) \cite{Neveol2016-ww}. Annotations in \quaero are based on the UMLS (2014AB version), restricted to ten semantic groups. Training, validation and test splits are provided, where the validation set of the 2016 dataset was the test set of the previous version (CLEF eHealth 2015 Task 1b). Annotation in \quaero has been carried out in a comprehensive fashion: nested entity spans are possible, and a single entity can be linked to multiple concept identifiers. The corpus provides 16,233 entity annotations, covering 5,130 unique CUIs.

The target KB resulting from the task-specific UMLS subset consists of 2.91M CUIs, for which we use 6.91M aliases in French and English. For \quaero, we initialize the BERT encoder of the fully supervised CE from the French biomedical BERT model DrBERT (4GB-CP-PubMedBERT) \cite{Labrak2023-qd}.

\subsection*{\bronco}

The German-language clinical corpus \bronco \cite{Kittner2021-cm} consists of 200 de-identified oncological discharge summaries, 150 of which are publicly available through a data use agreement (\textsc{Bronco150}). \bronco was annotated with three entity classes and concept identifiers from corresponding German versions of the following terminologies: ICD-10 (\textit{Diagnoses} annotations), OPS (\textit{Procedures}), and ATC (\textit{Medications}). 
In total, 4,080 mentions of diagnoses, 3,050 treatments, and 1,630 medications have been annotated. Within \textsc{Bronco150}, five pre-defined cross-validation folds are provided, which we use to evaluate our models.

For diagnoses and treatments, we follow the target KB definition from \citet{Kittner2021-cm} as closely as possible by using concepts and aliases from the German versions of ICD-10 and OPS, only. However, for medication names, the authors also report using the \textit{Rote Liste} of drug names \cite{roteliste}. As this is a commercial database, we instead obtain trade names from the free resource \textsc{DrugBank} (version 5.1.10) to obtain additional (English) aliases for ATC codes \cite{Wishart2017-tl}. The fully supervised CE model was initialized from the German biomedical model \textsc{BioGottBERT} \cite{Lentzen2022-ia}.

\subsection*{\distemist}

The \distemist shared task was part of the 10th \textsc{BioASQ} lab \cite{Miranda-Escalada2022-bv}. The dataset consists of Spanish-language clinical case reports from various medical specialties, annotated with disease mentions and a task-specific set of 111K SNOMED CT concept codes, provided as part of the \distemist gazetteer. In total, \distemist has 8,087 entity annotations linked to 3,297 unique CUIs. We use the official training and test splits provided by the task organizers. As a validation set for model selection, we use a subset of 20\% of the training set, which we have shown to be representative of the distribution of the held-out test set in earlier work \cite{Borchert2022-hn}. 

For the 111K concepts in the official gazetteer, we obtain 1.52M Spanish and English aliases through the UMLS metathesaurus (2022AA). Here, we consider only UMLS concepts which can be mapped to one the SNOMED CT codes in the \distemist gazetteer. The fully supervised CE is initialized from a Spanish biomedical-clinical RoBERTa model provided by \citet{Carrino2022-wk}.


\newpage
\section{Machine Translation and Entity Mapping}
\label{appendix:translate}
\autoref{tab:translation} shows the results of applying NMT and label projection to \medmentions. As expected, the entity alignment is imperfect, with up to 10.49\% of labels that could not be recovered after the mapping (usually because of syntax errors, e.g., missing start or end tags). The loss for German and Dutch is much smaller—one reason might be that these belong to the same language family as the source language. We report the final test $F_1$ scores after CE training on the translated datasets as a sanity check. Except for English, these are substantially below the reported state-of-the-art results on the English \medmentions dataset (75.73\% accuracy reported by \citet{Agarwal2022-fd}).

\begin{table}[ht]
\centering
    \caption{Automatically translated versions of \medmentions, number of entities after label projection, and the relative loss in the number of entities compared to the source dataset. Furthermore, we report the test set $F_1$ score of the CE model trained for five epochs on these weakly labeled datasets.}
    \label{tab:translation}
    \begin{tabular}{lrrr}
    \toprule
    Language &  \# Entities & Loss (\%) & CE $F_1$ \\
    \midrule
    English (original) & 203,282 & - & .722\\
    Dutch & 200,231 & 1.50 & .624 \\
    German & 199,006 & 2.10 & .598 \\
    Spanish & 185,029 & 8.98 & .569 \\
    French & 181,958 & 10.49 & .556 \\
\botrule
\end{tabular}
\end{table}

\end{document}